\documentclass[letterpaper]{article} 
\usepackage{aaai25}  
\usepackage{times}  
\usepackage{helvet}  
\usepackage{courier}  
\usepackage[hyphens]{url}  
\usepackage{graphicx} 
\usepackage{natbib}  
\usepackage{caption} 
\usepackage{algorithm}
\usepackage{algorithmic}

\usepackage{newfloat}
\usepackage{listings}
\DeclareCaptionStyle{ruled}{labelfont=normalfont,labelsep=colon,strut=off} 
\floatstyle{ruled}
\newfloat{listing}{tb}{lst}{}
\floatname{listing}{Listing}
\title{FaceCat: Enhancing Face Recognition Security with a Unified Diffusion Model}

\author {
    Jiawei Chen\textsuperscript{\rm 1},
    Xiao Yang\textsuperscript{\rm 2},
    Yinpeng Dong\textsuperscript{\rm 2},
    Hang Su\textsuperscript{\rm 2},
    Zhaoxia Yin\textsuperscript{\rm 1}
}
\affiliations {
    \textsuperscript{\rm 1}Shanghai Key Laboratory of Multidimensional Information Processing, East China Normal University\\
    \textsuperscript{\rm 2}Department of Computer Science \& Technology, Tsinghua University\\
}

\usepackage{bibentry}

\usepackage{multirow}
\usepackage{bbding}
\usepackage{amsmath}
\usepackage{subcaption}
\usepackage{amssymb}
\usepackage{bm}

\begin{document}

\maketitle

\begin{abstract}
Face anti-spoofing (FAS) and adversarial detection (FAD) have been regarded as critical technologies to ensure the safety of face recognition systems. However, due to limited practicality, complex deployment, and the additional computational overhead, it is necessary to implement both detection techniques within a unified framework. This paper aims to achieve this goal by
breaking through two primary obstacles: 1) the suboptimal face feature representation and 2) the scarcity of training data. To address the limited performance caused by existing feature representations, motivated by the rich structural and detailed features of face diffusion models, we propose \textbf{FaceCat}, the first approach leveraging the diffusion model to simultaneously enhance the performance of FAS and FAD. Specifically, FaceCat elaborately designs a hierarchical fusion mechanism to capture rich face semantic features of the diffusion model. These features then serve as a robust foundation for a lightweight head, designed to execute FAS and FAD simultaneously. Due to the limitations in feature representation that arise from relying solely on single-modality image data, we further propose a novel text-guided multi-modal alignment strategy that utilizes text prompts to enrich feature representation, thereby enhancing performance. To combat data scarcity, we build a comprehensive dataset with a wide range of 28 attack types, offering greater potential for a unified framework in facial security. Extensive experiments validate the effectiveness of FaceCat generalizes significantly better and obtains excellent robustness against common input transformations.
\end{abstract}

%

\section{Introduction}
\label{sec:intro}
Deep learning has propelled face recognition (FR) to the forefront of biometric applications, but its proliferation has sparked security concerns like presentation attacks~\shortcite{liu2019deep,oulu,hifimask} and adversarial attacks~\shortcite{yang2023towards,advhat,dong2019efficient}, which are crafted to deceive FR systems into granting unauthorized access or misidentifying individuals.
In response to these challenges, specialized defense mechanisms have been developed, namely face anti-spoofing~\shortcite{CDCN,frt,flexibleyu} and face adversarial detection~\shortcite{dfraa,libre}. Both of them are considered as two independent tasks by the existing methods. This suggests the necessity for deploying multiple models, thereby increasing the computational overhead. Moreover, recent research~\shortcite{deb2023unified,yu2022bench,chen2023advfas} indicates that these methods generally lack generalization ability (The FAS method cannot be directly applied to the detection of adversarial examples, and vice versa). In other words, these defenses cannot generalize well on unknown attack categories due to the overfitting to the manipulation types they are trained, limiting the practical applicability.
Therefore, it is critical for FR systems to integrate these security tasks to improve their overall robustness against attacks.
\begin{figure}[t]
  \centering
   \includegraphics[width=0.99\linewidth]{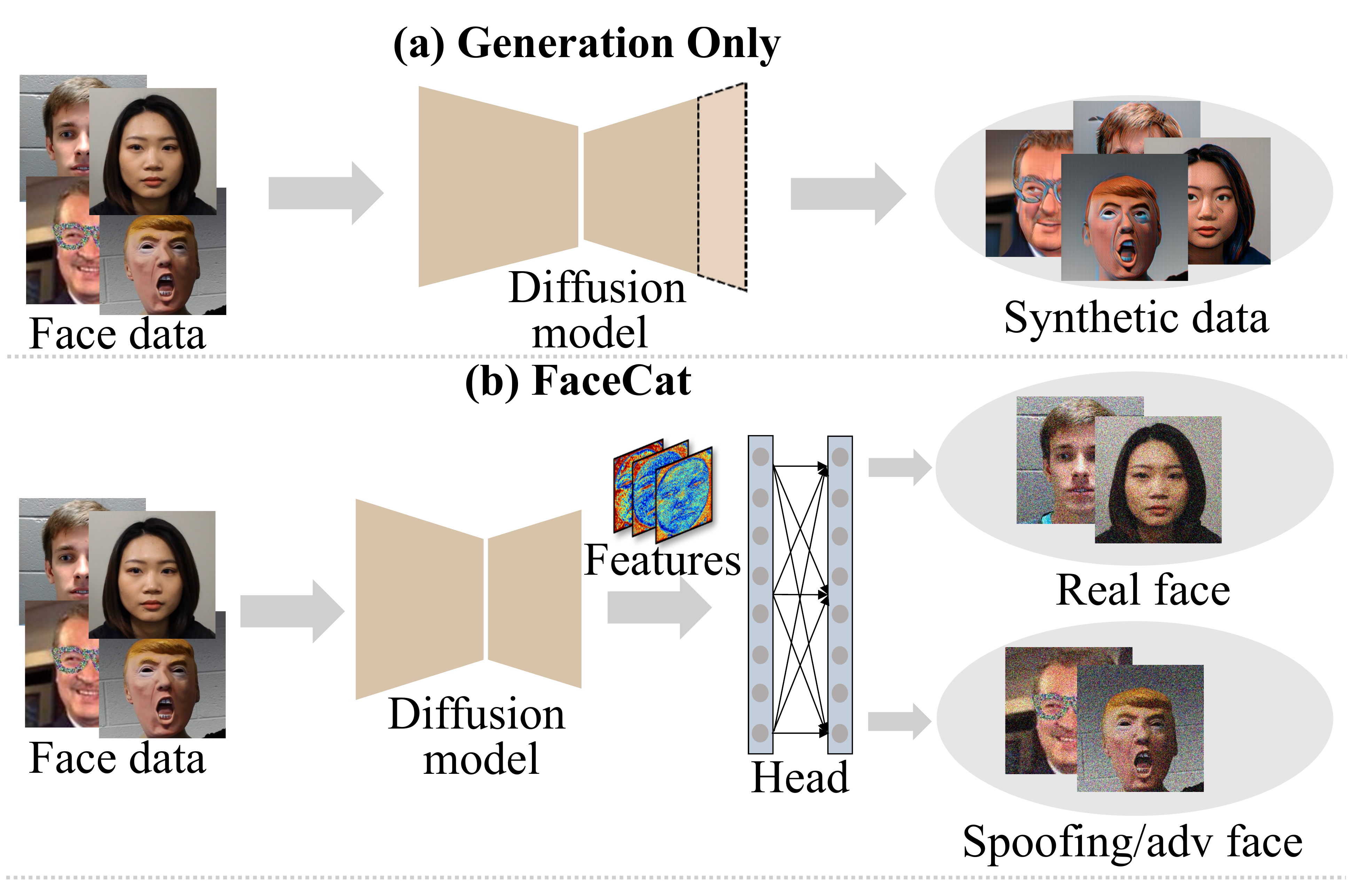}
   \caption{Comparison of our method and face generative models. (a) The face generation only models. (b) Our FaceCat exploits the abundant features inherent in face generative models to serve face anti-spoofing and adversarial detection simultaneously.}
   \label{fig: fig1}
\end{figure}

To address the problem mentioned above, there are two main challenges: (1) \textit{Suboptimal face feature representation}: traditional methods, which often treat FAS and FAD tasks as classification problems and utilize discriminative models to address multiple attack categories simultaneously~\shortcite{deb2023unified,yu2022bench}, are not entirely suitable for a unified facial security framework. Such approaches typically tend to make models primarily focus on the structural features of images~\shortcite{diffmae}—A richer representation of facial features is essential for the unified facial security framework, where our aim is not only adversarial examples but also spoofing samples. Furthermore, these approaches rely solely on single-mode facial data, which limits their ability to further express features~\shortcite{flexibleyu}. (2) \textit{Data scarcity}: acquiring facial training data that covers a wide range of attack types is challenging compared to traditional facial data. Although some facial security datasets have been introduced, they exhibit significant shortcomings in terms of data types, such as the lack of 3D data and adversarial examples. This limitation hinders the development of facial security frameworks.

\begin{figure*}[t]
\begin{center}
\includegraphics[width=0.99\linewidth]{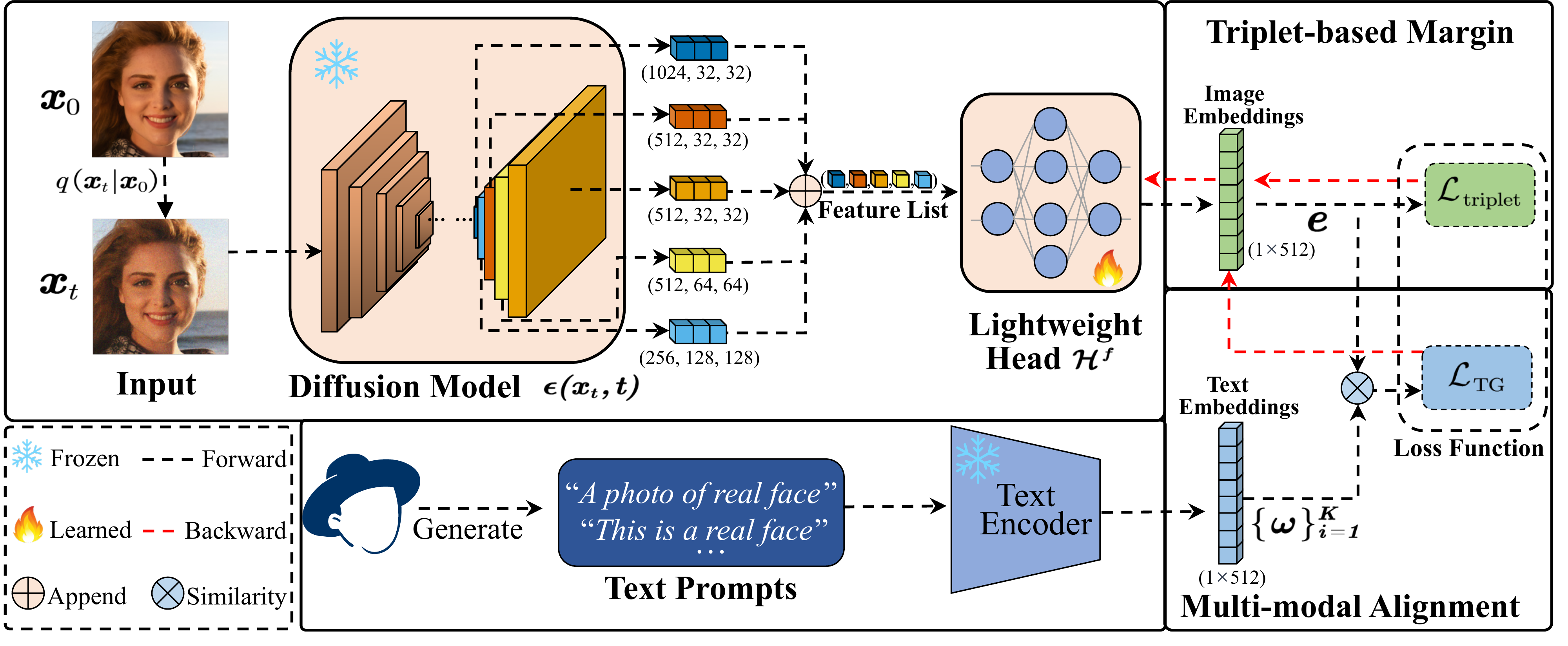}
\end{center}
\caption{An overview of our proposed FaceCat framework. FaceCat includes a generative model $\epsilon(\bm{x}_t,t)$ to encode the noisy image $\bm{x}_t$ into the face features, a lightweight head $\mathcal{H}^f$ to extract image embeddings $e$ from these face features, and a text encoder to obtain text embeddings from text prompts. Through image embeddings and text embeddings $\{\bm{\omega}\}_{i=1}^K$, the multi-modal alignment strategy calculates a text-image similarity score treated as the logit. The triplet-based margin is utilized to facilitate the learning of features.} 
\label{fig:frame}
\end{figure*}

Based on the above discussions, this paper conducts a pioneering exploration of the unified facial security framework. Specifically, we address the aforementioned challenges by making the following contributions:

\begin{table}[t]
\scriptsize
\setlength\tabcolsep{3.0pt}
\renewcommand\arraystretch{1.0}
\centering

\begin{tabular}{c|cccccc}
\hline
\multirow{2}{*}{Study}  & \multicolumn{2}{c|}{Presentation attack} & \multicolumn{3}{c|}{Adversarial attack}& \multirow{2}{*}{Types}
\\ \cline{2-6}
&\multicolumn{1}{c}{~~~~~2D}  & \multicolumn{1}{c|}{3D} & \multicolumn{1}{c}{perturbation} & \multicolumn{1}{c}{patch} &\multicolumn{1}{c|}{3D-printed}
\\ \hline
FaceGuard~\shortcite{faceguard}    &~~~~~ \XSolidBrush   & \multicolumn{1}{c|}{\XSolidBrush } 
             & \Checkmark  & \XSolidBrush & \multicolumn{1}{c|}{\XSolidBrush} & 6 \\
SIM-Wv2~\shortcite{liu2019deep}      &~~~~~ \Checkmark & \multicolumn{1}{c|}{\Checkmark} &  \XSolidBrush 
            & \XSolidBrush &  \multicolumn{1}{c|}{\XSolidBrush}   & 14\\
WMCA~\shortcite{wmca}        &~~~~~ \Checkmark & \multicolumn{1}{c|}{\Checkmark} &  \XSolidBrush   
             & \XSolidBrush & \multicolumn{1}{c|}{\XSolidBrush} & 7          \\
HQ-WMCA~\shortcite{hqwmca}      &~~~~~ \Checkmark  & \multicolumn{1}{c|}{\Checkmark} &  \XSolidBrush 
             & \XSolidBrush &  \multicolumn{1}{c|}{\XSolidBrush}   & 10          \\
HiFiMask~\shortcite{hifimask}     &~~~~~ \XSolidBrush & \multicolumn{1}{c|}{\Checkmark} & \XSolidBrush  
                 & \XSolidBrush & \multicolumn{1}{c|}{\XSolidBrush} & 3        \\
GrandFake~\shortcite{deb2023unified}    &~~~~~  \Checkmark &\multicolumn{1}{c|}{ \Checkmark} &  \Checkmark
            & \XSolidBrush & \multicolumn{1}{c|}{\XSolidBrush} & 25\\ \hline
\textbf{FaceCat (Ours)} &~~~~~ \Checkmark& \multicolumn{1}{c|}{\Checkmark}                                    & \Checkmark & \Checkmark & \multicolumn{1}{c|}{\Checkmark}&\textbf{28}\\ \hline
\end{tabular}
\caption{Comparison of the proposed protocol with others.}
\label{table: dataset comparison}
\end{table}
\textit{Effective framework for integrating facial security tasks}. We propose \textbf{FaceCat}, a novel framework to integrate FAS and FAD tasks.  As illustrated in Figure~\ref{fig:frame}, FaceCat extracts face features by leveraging a face diffusion model (FDM)~\shortcite{ddpm}, which offers a wealth of facial features (as shown in Figure~\ref{fig: feature}) for enhanced facial security tasks. To address the challenge of adapting FDM to face security tasks, we design a hierarchical fusion mechanism to extract the incorporation of multi-level features for rich face representations. Specifically, we amalgamate five distinct hierarchical blocks and stack them through pooling operations to form a unified face feature, adept at seamless integration with downstream network architectures.
Moreover, To address the limitations of single-modal data, we propose a text-guided (TG) multi-modal alignment strategy, which utilizes text knowledge to enrich semantic content and thereby further enhance the FaceCat's representation capability. In detail, we introduce textual information to describe visual concepts and generate text embeddings. These embeddings are then integrated with FDM's image embeddings to compute a similarity score, which is regarded as the logit for optimization.
Besides, owing to the similarity of face features, a triplet-based margin optimization is adopted to make real samples more compact and push attack samples apart.
 
\textit{Thorough facial security dataset.} We introduce \textbf{FaceCatData}, a comprehensive and fair dataset (approximately 410,000 face images) tailored for unified face security tasks. In particular, this dataset encompasses \textbf{28 diverse attack types}, involving multiple new practical attacks, such as 3D-printed attacks, as shown in Table \ref{table: dataset comparison}.
Based on this dataset, we compare FaceCat with several currently popular methods, including \textbf{five} FAS methods, \textbf{three} FAD methods, and \textbf{four} classification models. For the sake of fairness, all these methods have been fine-tuned on the proposed protocol. Moreover, three prevalent input transformations~\shortcite{jpeg,blur} are adopted to verify the robustness of FaceCat. We also conduct the ablation study to further investigate the proposed components. Experimental results demonstrate that our method exhibits excellent performance and outstanding robustness. Our main contributions can be summarized as: 
\begin{itemize}
    \item To the best of our knowledge, this is the first work to unify FAD and FAS tasks using a diffusion model to serve as a strong feature initialization.
    \item We propose a text-guided multi-modal alignment strategy with text embeddings which can enhance performance by leveraging multi-modal signals' supervision.
    \item We develop a comprehensive and fair dataset to perform FAS and FAD simultaneously. 
    \item Extensive experiments demonstrate that the proposed method exhibits superior performance in the task of FAD and FAS.
\end{itemize}

\section{Related Work}
In this section, we review related work on representation with generative models, FAS, and FAD. 

\subsection{Representation with Generative Models}
Pioneers attempt to exploit GANs~\shortcite{gan1} and VAEs~\shortcite{vae1} for representation learning.~\shortcite{bigbigan} demonstrates GANs can learn a competitive representation for images in latent space. Over the recent biennium, diffusion models~\shortcite{ddpm,ddpm-segmentation} have garnered substantial achievements in the realm of generative learning. Therefore, recent works attempt to utilize diffusion models for representation learning to serve downstream tasks, such as image editing~\shortcite{diffusionclip}, semantic segmentation~\shortcite{ddpm-segmentation} etc. Despite achieving commendable performance, the majority of these studies lack investigation into integrating face security tasks and the robustness of diffusion models.

\subsection{Face Anti-Spoofing and Adversarial Detection}

FAS~\shortcite{frt,srivatsan2023flip,chen2023advfas} and FAD~\shortcite{dfraa, fds, est, libre} have been developed to protect face recognition systems. In recent years, research in FAS has increasingly gravitated towards the exploration of 3D and multi-modal approaches~\shortcite{hqwmca}. FAD exhibits a preference for the development of attack-agnostic, universally applicable defense methods~\shortcite{libre,attack-agnostic}. Moreover, some research~\shortcite{deb2023unified,yu2022bench} attempts to unify several face security tasks into a framework, aiming to enhance both practicality and generalization. UniFAD~\shortcite{deb2023unified} attempts to unify digital attacks and FAS  through k-means clustering.  However, due to the inherent drawback of classification models, such methods primarily focus on the structural features of images. As a comparison, FaceCat has global structural and deep-level detailed features by virtue of its use of generative models.

\section{Method}
We first formulate how to unify FAS and FAD. Afterward, we elucidate the technical details of the hierarchical fusion mechanism and the lightweight head. We then propose a text-guided multi-modal alignment strategy for improving performance. Moreover, a triplet-based margin optimization is utilized to auxiliarily increase the distance between positive and negative samples. An overview of our proposed method is provided in Figure \ref{fig:frame}.

\subsection{Problem Formulation}
\label{sec:3.1}

For FAS and FAD tasks, we can define three input types: spoofing face $\bm{x}^{spoof}$, adversarial face $\bm{x}^{adv}$ and real face $\bm{x}^{real}$.
As for a unified detector,
it is only necessary to reject potential attack types without identifying whether $\bm{x}^{spoof}$ or $\bm{x}^{adv}$. Thus, we define $\bm{x}^{fake} \in \left\{ \bm{x}^{spoof}, \bm{x}^{adv} \right\}$. 
The decision function can be defined as:
\begin{align}
    f_{\bm{\theta}}: \bm{x} \rightarrow \{0 \text{ ($\bm{x}^{fake}$)}, 1 \text{ ($\bm{x}^{real}$)}\}, \nonumber
\end{align}
which classifies whether a sample $\bm{x}$ is fake or real. $f_{\bm{\theta}}(\bm{x}):\mathbb{R}^d \rightarrow \mathbb{R}$ is the unified detector parameterized by $\bm{\theta}$.
Generally, the problem is formulated as a dichotomous classification as follows:
\begin{small}
\begin{gather}
    \mathcal{L}_{c}(\bm{x},y) =\mathbb{E}_{p(\bm{x})}\left[-(y\log (f_{\bm{\theta}}(\bm{x}))+(1-y)\log (1-f_{\bm{\theta}}(\bm{x})))\right], \nonumber\\
    \text{where }y \in \left\{0,1\right\}, \bm{x} \in \left\{\bm{x}^{fake}, \bm{x}^{real}\right\},
        \label{form: first}
\end{gather}
\end{small}where $y$ is the ground truth, $\bm{x} \in \mathbb{R}^d$ is the input. 
In this paper, our method also considers unifying FAS and FAD as a dichotomous classification problem.

\begin{figure*}[t]
  \centering
   \includegraphics[width=0.99\linewidth]{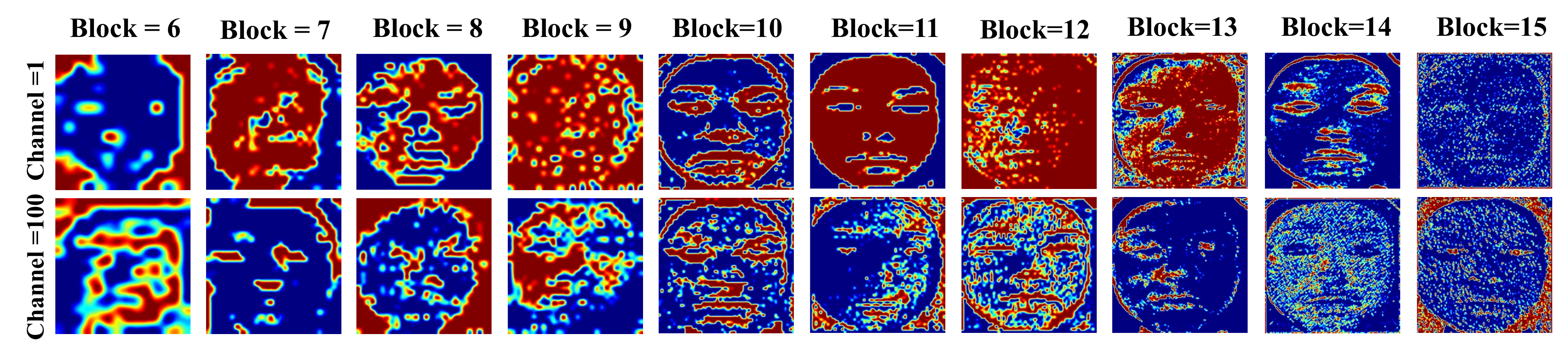}
   \caption{Multi-block feature representations are derived from the face diffusion model across different channels.}
   \label{fig: feature}
\end{figure*}
Traditional classification models may overly prioritize image structure at the expense of fine details, whereas FDM inherently contains rich global structural and deep-level detail features. Therefore, we aim to leverage it as a strong initialization for FAS and FAD.
A brief overview of FDM reveals that its training process is structured in two stages: 1) The forward diffusion process gradually adds Gaussian noise to the data to obtain a sequence of noisy samples; 2) The backward generative process reverses the diffusion process to denoise images. The central objective is to predict the noise at time $t$, which is formulated as a simple mean squared error loss:
\begin{equation}
\label{eq:diffusion-loss}
\begin{split}
\mathcal{L}_{\text{simple}}=\mathbb{E}_{\bm{x}_0,t,\bm{\epsilon}}[\|\epsilon_{\bm{\psi}}(\bm{x}_t, t) - \bm{\epsilon} \|_2^2]
\end{split},
\end{equation}
where $\epsilon_{\bm{\psi}}$ is the diffusion models and $\bm{x}_t$ is the noisy image at time $t$ (noise level), i.e., $\bm{x}_t$ is obtained by adding noise $\epsilon$ to $\bm{x}$. 

In this paper, we focus on using the \textbf{Face} diffusion model as a \textbf{Cat}alyst (\textbf{FaceCat}) for enhancing FAS and FAD. Formally, given a FDM $\epsilon_{\bm{\psi}}$, Note that we only obtain the features from the middle layer through $\epsilon_{\bm{\psi}}$, not the predicted noise, i.e., the input $\bm{x}_t$ is processed through $\epsilon_{\bm{\psi}}$ to obtain a universal face feature with dimension $d'$. To leverage this feature, a lightweight head $\mathcal{H}: \mathcal{R}^{d'} \rightarrow \mathcal{R}$ is adopt. Therefore, the objective function of FaceCat can be formulated as:
\begin{gather}
    \underset{\bm{\kappa}}{\min}~\mathbb{E}_{p(\bm{x}_t)}\left[\mathcal{L}_{c}(\mathcal{H}_{\bm{\kappa}}(\epsilon_{\bm{\psi}}(\bm{x}_t,t)),y)\right],
        \label{form: second}
\end{gather}
where $\mathcal{H}$ is parameterized by $\bm{\kappa}$ and $p(\bm{x}_{t})$ is the probability distribution of $\bm{x}_t$. Note that Equation \ref{form: second} is only adopted to optimize the lightweight head $\mathcal{H}_{\bm{\kappa}}$. The rationale behind this is twofold: 1) The features of FDM are sufficiently general-purpose; 2) By not fine-tuning FDM, we reduce the training time. By following the objective \eqref{form: second}, given a face image $\bm{x}$, we can get the probability of whether it is a real face.

\subsection{Hierarchical Fusion Mechanism and Lightweight Head}
\label{sec:3.4}

\begin{table}[t]
    \begin{center}
    \begin{tabular}{c|ccc}
        \hline
        \multirow{2}{*}{}& 
        \multicolumn{3}{c}{Performance}\\
        \cline{2-4}
                 &ACER &EER & TDR@0.2\%FDR \\
        \hline
        B = 5, 6&2.62&2.59&92.81\\
        B = 7, 8&2.67&2.74&92.89\\
        B = 5, 12&2.82&2.79&92.58\\
        B = 5, 6, 7, 12&2.56&2.40&93.07\\
        B = 5, 6, 7, 8&2.51&2.45&93.19\\
        B = 5, 6, 7, 8, 12&\textbf{1.36}&\textbf{2.12}&\textbf{94.64}\\
        \hline
    \end{tabular}
    \end{center}
    \caption{The performance of FaceCat with different blocks.}
    \label{tab:block}
\end{table} 
As illustrated in Figure \ref{fig: feature}, different blocks of the diffusion model capture distinct facial features, with some layers representing detailed features and others abstract features. For facial security tasks, both types of features are beneficial to model performance. Thus, we propose leveraging a hierarchical fusion mechanism to integrate these diverse features:
\begin{equation}
    f_{total} = \sum_{i=1}^{n} \mathcal{F}_i(f_i),
\end{equation}
where $f_i$ denotes the input features to the $i^{th}$ block, and $\mathcal{F}_i(f_i)$ represents the output features after processing by the $i^{th}$ block. $f_{total}$ is the features after fusion. For the lightweight head, aiming for minimal weight while ensuring performance, we utilize the first four layers of ResNet-18, complemented by an additional two downsampling layers.

Although some related works~\shortcite{depthnet} have also introduced the concept of multi-level features, their application has largely been empirical within traditional classification models. Given the unique architecture of diffusion models, which differs significantly from conventional classification models, previous experiences cannot be directly applied. Consequently, our investigation into the fusion of different blocks is detailed in Table \ref{tab:block}. Observations indicate that optimal performance is achieved when blocks 5, 6, 7, 8, 12 are selected. Consequently, this configuration has been adopted for subsequent experiments.
\subsection{Text-guided Multi-modal Alignment}
\label{sec:3.2}
The performance of FAS and FAD often suffers due to the constraints associated with the exclusive use of single-modal (RGB) data. Drawing inspiration from the ability of textual data to impart more detailed information in computer vision tasks, we try to design a text-guided multi-modal alignment strategy so that text information can be used in face security models.

Our text-guided multi-modal alignment is based on CLIP~\shortcite{clip}. Given a batch of image-text pairs, we can predict the image's class by computing a text-image cosine similarity. Formally, given an image $\bm{x}$, we can compute the prediction probability as follows:
\begin{equation}
\label{form:three}
p(\hat{y}|\bm{x}) = \frac{\text{exp}(\text{cos}(\bm{\omega}_i, \bm{e})/\tau)}{\sum_{j=1}^{K}\text{exp}(\text{cos}(\bm{\omega}_i, \bm{e})/\tau)},\end{equation}
where $\tau$ is a temperature parameter and $\hat{y}$ is the
predicted class label. $\bm{e}$ is the image embedding of $\bm{x}$ extracted by the image encoder, and \{$\bm{\omega}\}_{i=1}^K$ represent the weight vectors from the text encoder. $K$ and cos($\cdot,\cdot$) denote the number of classes and cosine similarity, respectively.

\begin{table}[t]
    \begin{center}
    \begin{tabular}{c|ccc}
        \hline
        \multirow{1}{*}{}& 
        \multicolumn{1}{c}{\textbf{Text Prompts}}\\
        \hline
        \multirow{6}{*}{Real Prompts} 
                                 &A photo of a real face.\\
                                 &This is a real face.\\
                                 &This is not a fake face.\\
                                 &An example of a real face.\\
                                 &A photo of the bonafide face.\\
                                 &An example of a bonafide face.\\
        \hline
        \multirow{6}{*}{Fake Prompts} 
                                 &A photo of a fake face.\\
                                 &This is a fake face.\\
                                 &This is not a real face.\\
                                 &An example of a fake face.\\
                                 &A photo of the attack face.\\
                                 &An example of attack face.\\
        \hline
    \end{tabular}
    \end{center}
        \caption{The text prompts of the real and fake classes.}
      \label{tab:supp-tab1}
\end{table}
Specifically, we craft some text prompts for this task along the lines of ``a photo of a [class]". The ``[class]" is ``[real face]" or ``[fake face]", all text prompts are presented in Table \ref{tab:supp-tab1}. This can be formalized as $\bm{t} = [V]_1[V]_2\ldots[V]_M[\text{real/fake face}]$, where each $[V]_m (m \in \{1,\ldots,M\})$ is a vector with the same dimension as word embeddings (i.e., 512 for the text encoder), and $M$ is a hyperparameter specifying the number of context tokens. Given the text encoder $g(\cdot)$, we can obtain $\bm{\omega}=g(\bm{t})$.

As for $\bm{e}$, we can acquire through the lightweight head $\mathcal{H}$. 
Assume that the feature layer of $\mathcal{H}$ is $\mathcal{H}^f: \mathcal{R}^{d'} \rightarrow \mathcal{R}^{d^{\prime\prime}}$, where $d^{\prime\prime}=512$ which matches the dimension of word embeddings. Hence, Equation \ref{form:three} can be re-expressed as:
\begin{equation}
\label{form:four}
p(\hat{y}|\bm{x}_t) = \frac{\text{exp}(\text{cos}(g(\bm{t}), \mathcal{H}^{f}({\epsilon_{\bm{\psi}}}(\bm{x}_t)))/\tau)}{\sum_{j=1}^{K}\text{exp}(\text{cos}(g(\bm{t}), \mathcal{H}^{f}({\epsilon_{\bm{\psi}}}(\bm{x}_t)))/\tau)},
\end{equation}
where $K=2$ and $\bm{x}_t$ is the noisy image. Additionally, we employ the more flexible focal loss~\shortcite{lin2017focal} as the optimization function instead of binary cross-entropy loss. Formally, given that $p$ denotes the probability in $p(\hat{y}|\bm{x}_t)$ that the identified image is a real face, we adopt the notation $p_{c}$ to represent the probability of the target class:
\begin{equation}
\label{form:pc}
p_c = 
\begin{cases} 
p & \text{if } y = 1, \\
1-p & \text{otherwise.}
\end{cases}
\end{equation}
Referring to the standard $\alpha$-balanced focal loss~\shortcite{lin2017focal}, the text-guided multi-modal alignment can be formulated as follows:
\begin{equation}
\text{TG}(p_c,y) = -\alpha_c (1 - p_c)^\gamma \log(p_c),
\end{equation}
the parameter $\alpha_c$ is introduced to balance the data distribution, and its formal definition mirrors that of Equation \ref{form:pc}, with $p$ replaced by $\alpha$. Since negative samples are approximately twice as many as positive samples, $\alpha$ is set to 0.75. To encourage $\mathcal{H}^f$ to focus on hard samples during training, we set the focusing parameter $\gamma$ to 2.0.

\subsection{Triplet-based Margin Optimization}
\label{sec:3.3}
Face data from different classes exhibits similar features, which is different from general classification tasks. This characteristic makes it difficult to distinguish between positive and negative sample features. Therefore, we adopt a triplet loss~\shortcite{triplet} to supervise features, which is commonly utilized in deep metric learning, and has proven its effectiveness in optimizing the relative distances between samples in the embedded space. The mathematical form is expressed as follows:
\begin{equation}
\begin{split}
\mathcal{L}_{\text{triplet}}(\bm{x}^a, \bm{x}^p, \bm{x}^n) &= \max \left( 0, \left\| \mathcal{H}^f({\epsilon_{\bm{\psi}}}(\bm{x}^a)) - \mathcal{H}^f({\epsilon_{\bm{\psi}}}(\bm{x}^p))\right\|_2^2 \right. \nonumber \\
&\left. - \left\| \mathcal{H}^f({\epsilon_{\bm{\psi}}}(\bm{x}^a)) - \mathcal{H}^f({\epsilon_{\bm{\psi}}}(\bm{x}^n)) \right\|_2^2 + m \right),
\end{split}
\end{equation}
where $\bm{x}^p$ and $\bm{x}^n$ are the positive sample ($\bm{x}^{real}$) and negative samples ($\bm{x}^{fake}$) respectively. 
$\bm{x}^a$ is an anchor sample, which is the same class as $\bm{x}^{real}$. $\left\| \cdot \right\|_2^2$ represents the squared Euclidean distance. The margin $m$ is a hyperparameter ensuring a safe gap between positive and negative pairs, preventing trivial solutions. Therefore, Equation \ref{form: second} can be rewritten as:
{\small
\begin{gather}
    \underset{\bm{\kappa}'}{\min}~\mathbb{E}_{p(\bm{x}_t)}\left[\text{TG}(p_c,y)+\lambda \cdot \mathcal{L}_{triplet}(\bm{x}_t^a,\bm{x}_t^{real},\bm{x}_t^{fake})\right],
        \label{form: fomulation}
\end{gather}
}where $\bm{\kappa}'$ denotes the parameters of $\mathcal{H}^f$. $\lambda$ is the balancing factor. $\bm{x}_t^a$,$\bm{x}_t^{real}$ and $\bm{x}_t^{fake}$ represent the forms of $\bm{x}^a, \bm{x}^{real}$ and $\bm{x}^{fake}$ with noises. Note that we also do not optimize the parameters of $g(\cdot)$. Following Equation \ref{form: fomulation}, we can acquire a classifier that performs FAS and FAD simultaneously. 

\begin{table*}[t]
\setlength\tabcolsep{7pt}
\renewcommand\arraystretch{1.0}
\centering
\begin{tabular}{c|c|ccccc}
\hline
\multirow{2}{*}{}& \multirow{2}{*}{Method} & \multicolumn{5}{c}{Performance}   \\ \cline{3-7} 
     & & \multicolumn{1}{c|}{APCER} & \multicolumn{1}{c|}{BPCER} & ACER       
     & \multicolumn{1}{c|}{EER} &  TDR @ 0.2\% FDR       
     \\ \hline \hline
\multirow{5}{*}{Face anti-spoofing}
&DeepPixBis~\shortcite{deeppixbis}& \multicolumn{1}{c|}{1.89}& \multicolumn{1}{c|}{3.63}& \multicolumn{1}{c|}{2.76}& \multicolumn{1}{c|}{3.12}& 85.48           \\
&Depthnet~\shortcite{depthnet}& \multicolumn{1}{c|}{2.64}& \multicolumn{1}{c|}{5.53}& \multicolumn{1}{c|}{4.09}& \multicolumn{1}{c|}{4.23}& 88.81          \\
&FRT-PAD~\shortcite{frt}& \multicolumn{1}{c|}{2.80}  & \multicolumn{1}{c|}{3.04}& \multicolumn{1}{c|}{2.52}& \multicolumn{1}{c|}{2.93}& 89.20 
\\ 
& Flip~\shortcite{srivatsan2023flip}& \multicolumn{1}{c|}{1.82}& \multicolumn{1}{c|}{2.38}& \multicolumn{1}{c|}{2.10}& \multicolumn{1}{c|}{2.54}& 91.78         \\ 
& CFPL~\shortcite{cfplfas}& \multicolumn{1}{c|}{2.24}  & \multicolumn{1}{c|}{2.64}& \multicolumn{1}{c|}{2.44}& \multicolumn{1}{c|}{2.46}& 92.34\\ \hline
\multirow{3}{*}{Face adversarial detection} 
& EST~\shortcite{est}   & \multicolumn{1}{c|}{5.29} & \multicolumn{1}{c|}{2.01} &  \multicolumn{1}{c|}{3.65} & \multicolumn{1}{c|}{3.89}  & 88.31          \\
& FDS~\shortcite{fds}   & \multicolumn{1}{c|}{4.89} & \multicolumn{1}{c|}{3.22} & \multicolumn{1}{c|}{4.06} & \multicolumn{1}{c|}{3.54} & 87.43 \\
& DFRAA~\shortcite{dfraa} & \multicolumn{1}{c|}{3.52}  & \multicolumn{1}{c|}{3.12} & \multicolumn{1}{c|}{3.32} & \multicolumn{1}{c|}{3.26}  & 88.19  
         \\ \hline
\multirow{4}{*}{Classification models}
      & Resnet50~\shortcite{resnet}  & \multicolumn{1}{c|}{3.84} & \multicolumn{1}{c|}{2.77} 
            & \multicolumn{1}{c|}{3.31}& \multicolumn{1}{c|}{3.36}  & 86.41      \\
      & Inceptionv3~\shortcite{inception}   & \multicolumn{1}{c|}{2.91} & \multicolumn{1}{c|}{3.69} 
                 & \multicolumn{1}{c|}{3.30} & \multicolumn{1}{c|}{3.35} & 87.54 \\
    & Efficientnetb0~\shortcite{efficientnet}  & \multicolumn{1}{c|}{3.80}  & \multicolumn{1}{c|}{3.23} 
                & \multicolumn{1}{c|}{3.52} & \multicolumn{1}{c|}{3.53}  & 86.44 \\
      & ViT-b/16~\shortcite{vit} & \multicolumn{1}{c|}{3.31}  & \multicolumn{1}{c|}{3.02} 
                 &\multicolumn{1}{c|}{3.17} & \multicolumn{1}{c|}{3.18}  & 87.56   
                 \\ \hline
\multirow{2}{*}{\textbf{Proposed FaceCat}}
     & w/o TG   & \multicolumn{1}{c|}{1.96}  & \multicolumn{1}{c|}{2.70} & \multicolumn{1}{c|}{2.33} & \multicolumn{1}{c|}{2.35}  & 92.74 
     \\ 
    & with TG   & \multicolumn{1}{c|}{\textbf{1.36}}  & \multicolumn{1}{c|}{\textbf{2.12}} & \multicolumn{1}{c|}{\textbf{1.24}} & \multicolumn{1}{c|}{\textbf{1.68}}  & \textbf{94.64}   
         \\ \hline
\end{tabular}
\caption{The different performances comparison (\%) between the proposed FaceCat and baseline methods under the protocol.}
\label{table: big tab}
\end{table*}

\begin{figure}[t]
\begin{center}
\includegraphics[width=0.99\linewidth]{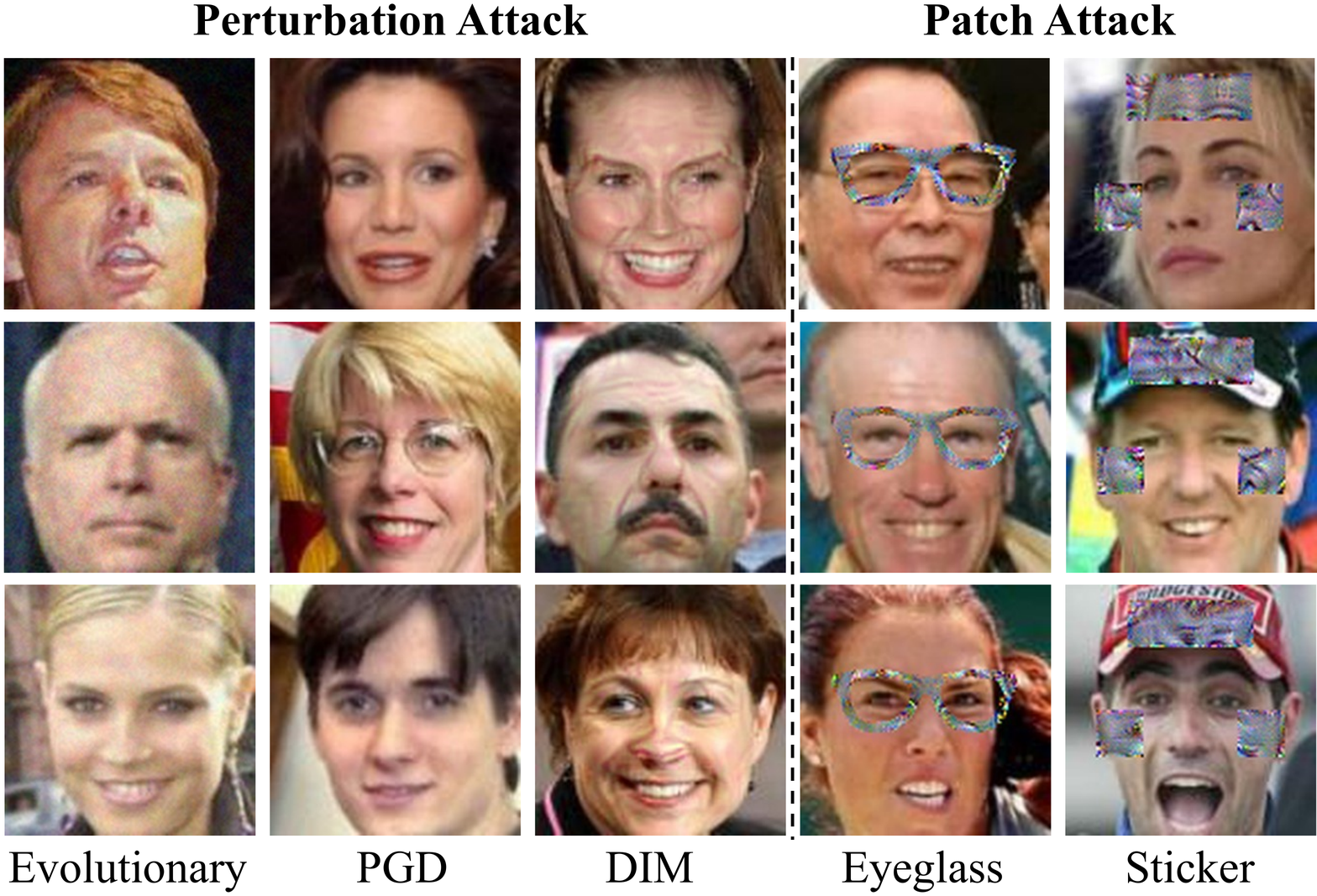}
\end{center}
\caption{The adversarial examples with different attacks.}
\label{fig: adversarial example}
\end{figure}

\section{FaceCatData}
\label{data}
\textbf{Dataset creation.}
To evaluate FaceCat's capacity for concurrent FAS and FAD, we first craft 14 types of adversarial examples from the LFW~\shortcite{lfw} dataset targeting the state-of-the-art (SOTA) face recognition model, ArcFace~\shortcite{arcface}, as shown in Figure \ref{fig: adversarial example}. We employ various techniques including: \textbf{1)} adversarial perturbation: FGSM(w)~\shortcite{fgsm}, PGD(w)~\shortcite{pgd}, DIM(w)~\shortcite{dim} and Evolutionary(b)~\shortcite{evolutionary}; \textbf{2)} adversarial patch~\shortcite{facesec}: Eyeglass(w), Sticker(w), Facemask(w). The `w' and `b' represent white and black-box attacks respectively. Each attack type contains dodging and impersonation attacks. All adversarial examples are generated using the default parameters of the respective work. The attack success rate (ASR) and more adversarial examples are shown in Appendix A. Moreover, SiW-Mv2 is adopted as a complement to evaluate FAS. We partition the adversarial examples based on Protocol 1 of SiW-Mv2 and subsequently incorporate them into SiW-Mv2. Therefore, we obtain a comprehensive dataset FaceCatData to evaluate FAS and FAD simultaneously.

\begin{table}[t]
    \begin{center}
    \begin{tabular}{c|c|ccc|c}
        \hline
        \multirow{2}{*}{Protocol}& \multirow{2}{*}{Class}& 
        \multicolumn{3}{c|}{Types}&\multirow{2}{*}{\# Total}\\
        \cline{3-5}
                 & &\# Live &\# Spoof &\# Adv \\ 
        \hline
        \multirow{3}{*}{protocol 1}&train&86404&68597&50000&205001\\
        &eval&34561&27379&20000&81940\\
        &test&51842&41218&30000&123060\\ \hline
                        \multirow{3}{*}{protocol 2}&train&86404&0&50000&136404\\
        &eval&34561&0&20000&54561\\
        &test&51842&41218&30000&123060\\ \hline
                \multirow{3}{*}{protocol 3}&train&86404&68597&0&155001\\
        &eval&34561&27379&0&61940\\
        &test&51842&41218&30000&123060\\ \hline
    \end{tabular}
    \end{center}
    \caption{The performance of FaceCat with different blocks.}
    \label{tab:protocol}
\end{table} 
\textbf{Types of protocols.} For the proposed FaceCatData, we design three different protocols, as shown in Table \ref{tab:protocol}. Protocol 1, known attack pattern detection: all attack types are present in the training, aiming to evaluate the method's capability to simultaneously perform FAS and FAD. Protocol 2 and Protocol 3 both involve unknown attack pattern detection. Specifically, the Protocol 2 is used to test the method's performance when spoof faces are absent from the training and validation sets. Protocol 3 assesses the method's performance when adversarial faces are not included in the training and validation sets. In this paper, unless stated otherwise, Protocol 1 is utilized and the experiments under Protocols 2 and 3 are detailed in Appendix B.

\section{Experiments}
In this section, we conduct extensive experiments to evaluate the performance of FaceCat and perform ablation studies to verify the effectiveness of each component.
\subsection{Experimental Settings}
\label{sec: 4.1}
\textbf{Evaluation metrics.}~We evaluate with the following metrics: Attack Presentation Classification Error Rate (APCER), Bona Fide Presentation Classification Error Rate (BPCER), and the average of APCER and BPCER, Average Classification Error Rate (ACER) for a fair comparison. Equal Error Rate (EER), Area Under Curve (AUC), and True Detection Rate (TDR) at a False Detection Rate (FDR) 0.2\% (TDR @ 0.2\% FDR). In addition, a visualization~\shortcite{tsne} is also reported to evaluate the performance further. The specific training details can be found in Appendix C.

\begin{figure*}[t]
  \centering
   \includegraphics[width=0.9\linewidth]{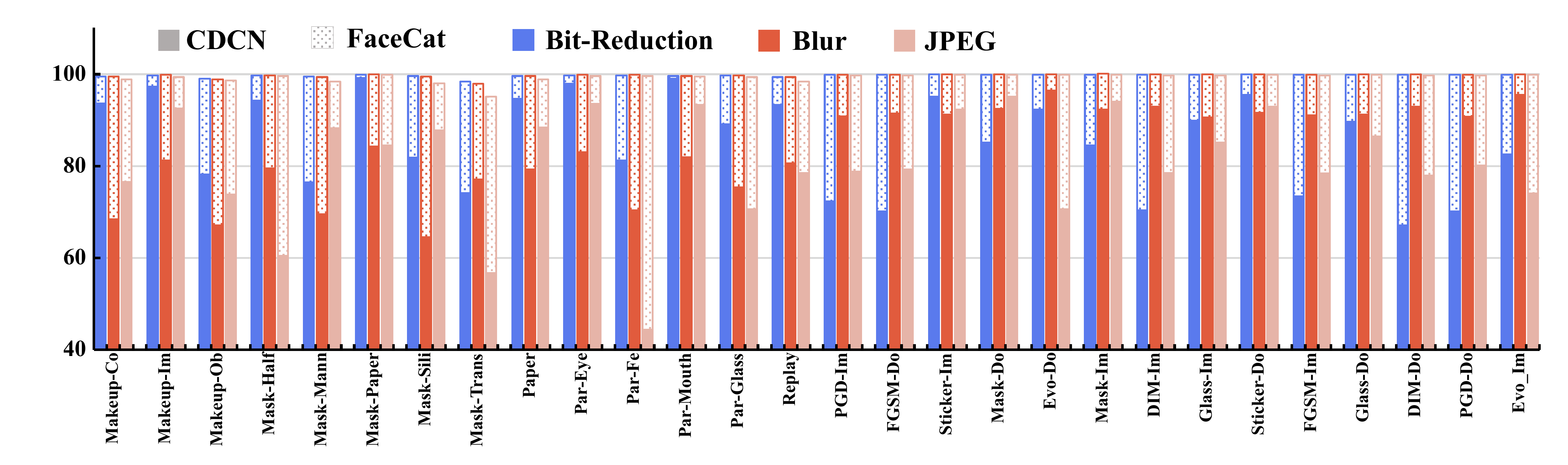}
   \caption{The area under curve (\%) between FaceCat and the baseline against common input transformations.}
   \label{fig: final}
\end{figure*}
\subsection{Experiment in the Proposed Protocol}
\textbf{Effectiveness of the proposed method.} To verify the efficacy of the proposed approach, we compare FaceCat with three baselines:~1)~FAS: Flip~\shortcite{srivatsan2023flip}, CFPL~\shortcite{cfplfas}, DeepPixBis~\shortcite{deeppixbis}, Depthnet~\shortcite{depthnet}, FRT-PAD~\shortcite{frt}; 2)~FAD: EST~\shortcite{est}, FDS~\shortcite{fds}, DFRAA~\shortcite{dfraa};~3)~classification models: Resnet50~\shortcite{resnet}, Inceptionv3~\shortcite{inception}, Efficientnetb0~\shortcite{efficientnet}, ViT-b/16~\shortcite{vit}. These methods are either commonly used or represent the SOTA techniques. To fairly compare the proposed method, all these methods have been fine-tuned according to the proposed protocol. Table \ref{table: big tab} shows the performance metrics for each method. Inspection of the table allows us to deduce the subsequent conclusions:

\textbf{\underline{(1)}} FaceCat effectively improves the face recognition systems' security. When conducting FAS and FAD simultaneously, all metrics achieve optimal performance, with ACER and EER being 1.24\% and 1.68\%, respectively. This can be attributed to: 1) the powerful representational capabilities of the diffusion model, which provide rich prior knowledge for FAS and FAD; 2) the proposed hierarchical fusion mechanism that effectively leverages the features of the diffusion model; 3) the text-guided multi-modal alignment technique that further enhances the performance of FaceCat by utilizing rich textual features.

\textbf{\underline{(2)}} The proposed method with the text-guided multi-modal alignment strategy generally outperforms the variant without it across various metrics. This demonstrates the strategy can effectively improve the performance of the proposed method.  We think it benefits from low-cost and effective text embedding, which facilitates face features learning under the nuanced supervision of multi-modal cues.

\textbf{Evaluation on input transformation.} To evaluate the robustness of FaceCat, common input transformations (are excluded from data augmentation), e.g., bit-depth reduction, Gaussian blur, and JPEG compression are employed to reduce the image quality. We compare our results with a widely recognized method, CDCN~\shortcite{CDCN}. In Figure \ref{fig: final}, when image quality degrades, the performance of CDCN in addressing diverse types of attacks diminishes, with its response to JPEG compression being the most notably affected. This indicates although the baseline can demonstrate decent performance under the proposed protocol, its robustness significantly decreases with the degradation of image quality. In contrast, FaceCat continues to maintain stable performance even in the face of degraded image quality, reaching up to 55.3\% better than the baseline under JPEG compression. This can be attributed to the superior feature robustness of FaceCat, i.e., even with the loss of certain image information, the proposed method is still sufficient to achieve robust metrics in detecting such attack threats.

\subsection{Ablation Study}
\begin{table}[t]
    \begin{center}
    \begin{tabular}{c|ccc}
        \hline
        \multirow{2}{*}{Method}& 
        \multicolumn{3}{c}{Performance}\\
        \cline{2-4}
                 &ACER &EER & TDR@0.2\%FDR \\
        \hline \hline
        MAE~\shortcite{mae}    &2.33&2.36&92.77\\
        SwAV~\shortcite{swav}   &2.36&2.23&91.57\\
        FR-ArcFace~\shortcite{arcface}     &2.75&2.41&91.14\\
        \hline
        w/o TG ($\alpha$=0.25)  &2.74&2.76&88.66\\
        w/o TG ($\alpha$=0.5)   &2.89&3.02&91.58\\
        w/o TG ($\alpha$=0.75)  &2.33&2.36&92.74\\ \hline
        \textbf{FaceCat}                   &\textbf{1.36}&\textbf{2.12}&\textbf{94.64}\\
        \hline
    \end{tabular}
        \caption{The ablation study of FaceCat in the protocol.}
          \label{tab: ablation}
    \end{center}
\end{table}

\textbf{Advantages of FDM architecture.} We conduct an ablation study to verify the effectiveness of FDM architecture, as shown in Table \ref{tab: ablation}. Both MAE~\shortcite{mae} and SwAV~\shortcite{swav} represent SOTA self-supervised learning methodologies that have undergone pre-training on the FFHQ dataset prior to fine-tuning on our proposed protocol, ensuring consistency in model initialization. The FR-ArcFace~\shortcite{arcface} is the face recognition model ArcFace, which utilizes a ResNet18~\shortcite{resnet} and is pre-trained for face recognition tasks. From the table, two key conclusions can be drawn: Firstly, even the employment of pre-trained face models can enhance the model's capabilities in conducting FAS and FAD; however, FDM, with its dual focus on global structural information and intricate detail features, demonstrates superior performance over these pre-trained face models. Secondly, utilizing face recognition models as pre-trained systems yields inferior results compared to models pre-trained via self-supervised methods. This is attributed to the tendency of classification models to overemphasize structural features, resulting in a partial loss of feature information.

\textbf{Value of $\bm{\alpha}$.} In Table \ref{tab: ablation}, we exploit the influence of different $\alpha$ of TG on the experimental results. $\alpha=0.75$ performs better than other settings. In joint defense scenarios, the quantity and diversity of real face samples are typically inferior to those of spoofing faces. This leads to an imbalanced focus of models on fake face samples. Hence, adjusting hyper-parameter $\alpha$ to mitigate sample imbalance significantly improves the model's overall performance.

\begin{figure}[t]
    \centering
    \scriptsize
    \includegraphics[width=\linewidth]{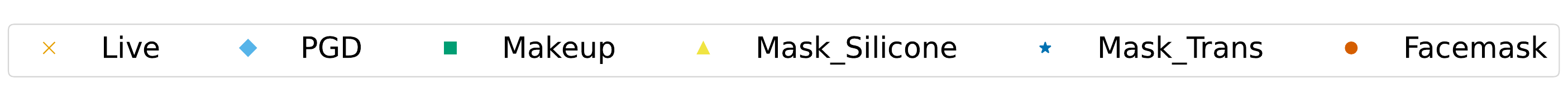}
    \begin{subfigure}{0.49\linewidth}
        \centering
        \includegraphics[width=\linewidth]{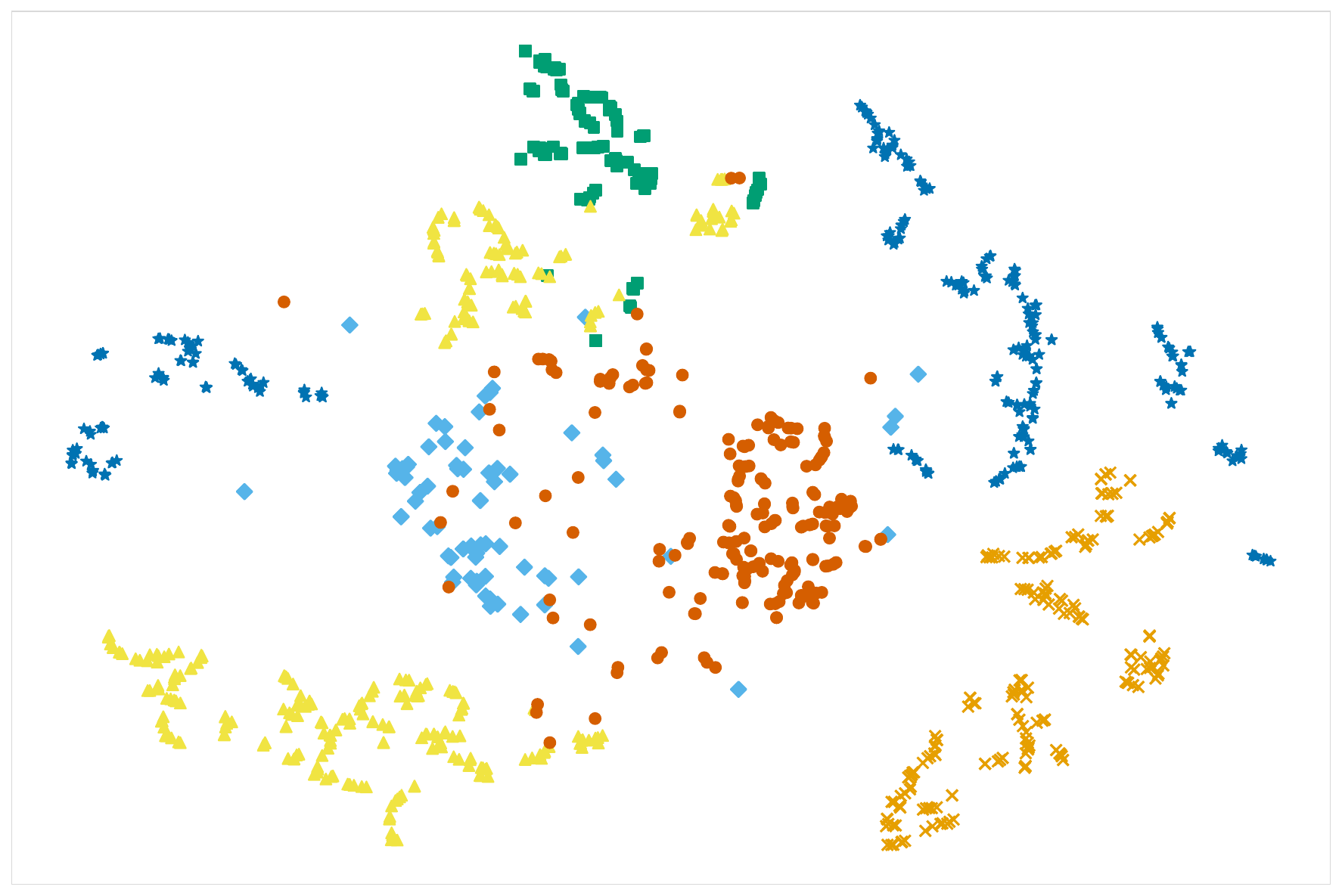}
        \caption{w/ TG}
        \label{fig:with-clip} 
    \end{subfigure}
    \hfill
    \begin{subfigure}{0.49\linewidth}
        \centering
        \includegraphics[width=\linewidth]{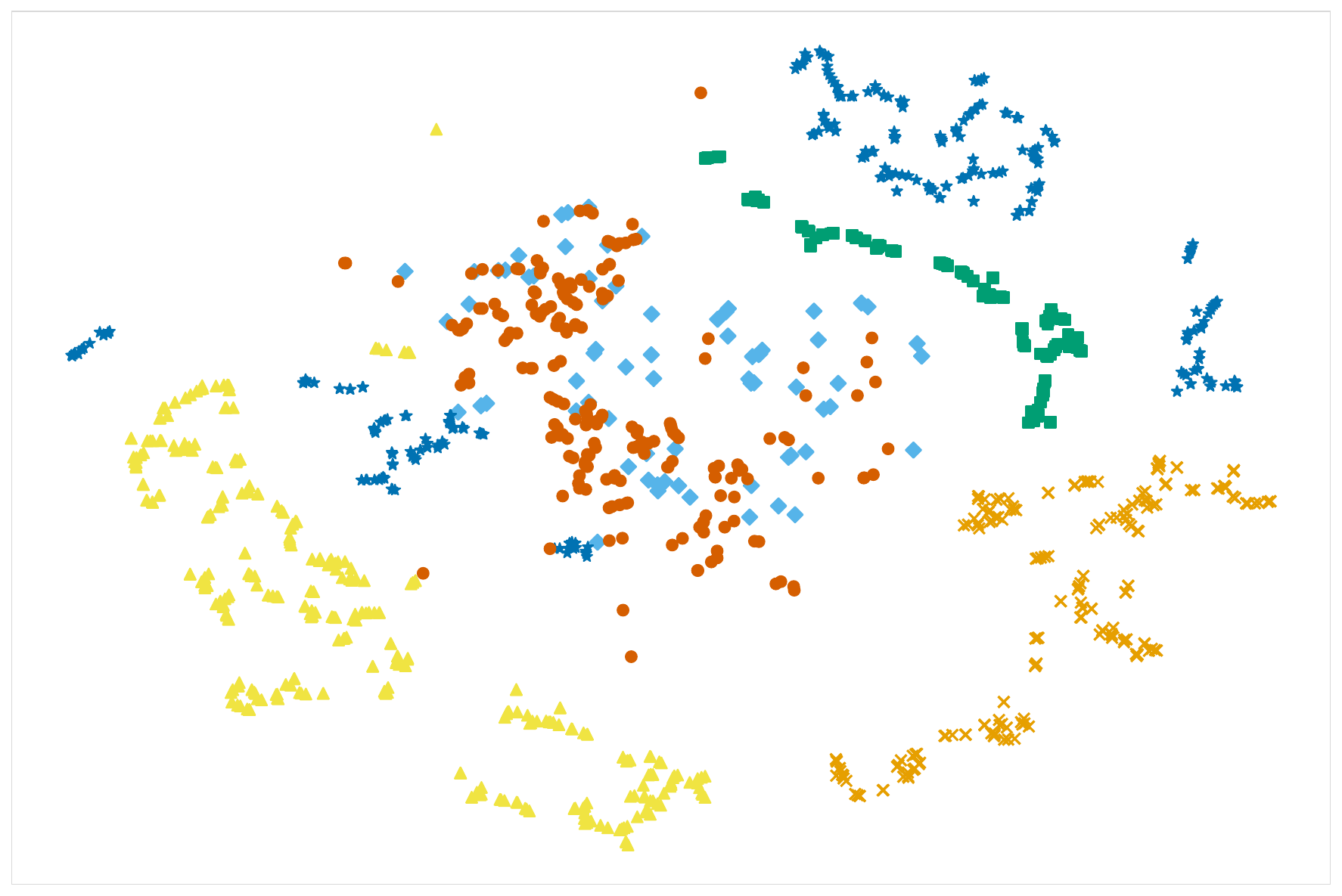}
        \caption{w/o TG}
        \label{fig:wo-clip} 
    \end{subfigure}
    \caption{Feature distribution visualization of the test set on the proposed protocol using t-SNE~\shortcite{tsne}. (a) Features with TG. (b) Features without TG.}
    \label{fig:feature_visualization}
\end{figure}

\textbf{Feature visualization.}~The feature distribution in the test set on the proposed protocol is visualized in Figure \ref{fig:feature_visualization} via t-SNE~\shortcite{tsne}, which consists of 6 types (live, PGD, makeup, silicone mask, facemask, transparent mask). Through the figure, it is evident from (b) that, in the absence of TG, facemask and PGD are not distinctly separable. Moreover, the live sample distribution from (a) is compact and the clusters exhibit improved separation for the proposed method trained with TG. These results illustrate the effectiveness of the text-guided strategy in FAS and FAD tasks. The main reason is that TG can leverage text embeddings to enhance the aggregation and distinction of image features.

The ablation study on the timesteps of diffusion and the real-world experiments are detailed in Appendix D and F.

\section{Conclusion}
In this paper, we proposed a novel framework called FaceCat, which treats the FDM as a pre-trained model for integrating face anti-spoofing and adversarial detection. Besides, we introduced a text-guided multi-modal alignment and hierarchical fusion mechanism to enhance semantic information and optimize the utilization of FDM's features, respectively. We also conducted extensive experiments to evaluate the effectiveness of FaceCat for face security tasks. Moreover, we further validated its robustness against common input transformations. 

\bibliography{aaai25}

\begin{thebibliography}{50}
\providecommand{\natexlab}[1]{#1}

\bibitem[{Baranchuk et~al.(2021)Baranchuk, Rubachev, Voynov, Khrulkov, and Babenko}]{ddpm-segmentation}
Baranchuk, D.; Rubachev, I.; Voynov, A.; Khrulkov, V.; and Babenko, A. 2021.
\newblock Label-efficient semantic segmentation with diffusion models.
\newblock \emph{arXiv preprint arXiv:2112.03126}.

\bibitem[{Boulkenafet et~al.(2017)Boulkenafet, Komulainen, Li, Feng, and Hadid}]{oulu}
Boulkenafet, Z.; Komulainen, J.; Li, L.; Feng, X.; and Hadid, A. 2017.
\newblock OULU-NPU: A mobile face presentation attack database with real-world variations.
\newblock In \emph{2017 12th IEEE international conference on automatic face \& gesture recognition (FG 2017)}, 612--618. IEEE.

\bibitem[{Caron et~al.(2020)Caron, Misra, Mairal, Goyal, Bojanowski, and Joulin}]{swav}
Caron, M.; Misra, I.; Mairal, J.; Goyal, P.; Bojanowski, P.; and Joulin, A. 2020.
\newblock Unsupervised learning of visual features by contrasting cluster assignments.
\newblock \emph{Advances in neural information processing systems}, 33: 9912--9924.

\bibitem[{Carrara et~al.(2018)Carrara, Becarelli, Caldelli, Falchi, and Amato}]{fds}
Carrara, F.; Becarelli, R.; Caldelli, R.; Falchi, F.; and Amato, G. 2018.
\newblock Adversarial examples detection in features distance spaces.
\newblock In \emph{Proceedings of the European conference on computer vision (ECCV) workshops}, 0--0.

\bibitem[{Chen et~al.(2023)Chen, Yang, Yin, Ma, Chen, Peng, Guo, Yin, and Su}]{chen2023advfas}
Chen, J.; Yang, X.; Yin, H.; Ma, M.; Chen, B.; Peng, J.; Guo, Y.; Yin, Z.; and Su, H. 2023.
\newblock AdvFAS: A robust face anti-spoofing framework against adversarial examples.
\newblock \emph{Computer Vision and Image Understanding}, 235: 103779.

\bibitem[{Deb, Liu, and Jain(2023{\natexlab{a}})}]{faceguard}
Deb, D.; Liu, X.; and Jain, A.~K. 2023{\natexlab{a}}.
\newblock Faceguard: A self-supervised defense against adversarial face images.
\newblock In \emph{2023 IEEE 17th International Conference on Automatic Face and Gesture Recognition (FG)}, 1--8. IEEE.

\bibitem[{Deb, Liu, and Jain(2023{\natexlab{b}})}]{deb2023unified}
Deb, D.; Liu, X.; and Jain, A.~K. 2023{\natexlab{b}}.
\newblock Unified detection of digital and physical face attacks.
\newblock In \emph{2023 IEEE 17th International Conference on Automatic Face and Gesture Recognition (FG)}, 1--8. IEEE.

\bibitem[{Deng et~al.(2019)Deng, Guo, Xue, and Zafeiriou}]{arcface}
Deng, J.; Guo, J.; Xue, N.; and Zafeiriou, S. 2019.
\newblock Arcface: Additive angular margin loss for deep face recognition.
\newblock In \emph{Proceedings of the IEEE/CVF conference on computer vision and pattern recognition}, 4690--4699.

\bibitem[{Deng et~al.(2021)Deng, Yang, Xu, Su, and Zhu}]{libre}
Deng, Z.; Yang, X.; Xu, S.; Su, H.; and Zhu, J. 2021.
\newblock Libre: A practical bayesian approach to adversarial detection.
\newblock In \emph{Proceedings of the IEEE/CVF conference on computer vision and pattern recognition}, 972--982.

\bibitem[{Donahue, Kr{\"a}henb{\"u}hl, and Darrell(2016)}]{gan1}
Donahue, J.; Kr{\"a}henb{\"u}hl, P.; and Darrell, T. 2016.
\newblock Adversarial feature learning.
\newblock \emph{arXiv preprint arXiv:1605.09782}.

\bibitem[{Donahue and Simonyan(2019)}]{bigbigan}
Donahue, J.; and Simonyan, K. 2019.
\newblock Large scale adversarial representation learning.
\newblock \emph{Advances in neural information processing systems}, 32.

\bibitem[{Dong et~al.(2019{\natexlab{a}})Dong, Su, Wu, Li, Liu, Zhang, and Zhu}]{dong2019efficient}
Dong, Y.; Su, H.; Wu, B.; Li, Z.; Liu, W.; Zhang, T.; and Zhu, J. 2019{\natexlab{a}}.
\newblock Efficient decision-based black-box adversarial attacks on face recognition.
\newblock In \emph{Proceedings of the IEEE/CVF Conference on Computer Vision and Pattern Recognition}, 7714--7722.

\bibitem[{Dong et~al.(2019{\natexlab{b}})Dong, Su, Wu, Li, Liu, Zhang, and Zhu}]{evolutionary}
Dong, Y.; Su, H.; Wu, B.; Li, Z.; Liu, W.; Zhang, T.; and Zhu, J. 2019{\natexlab{b}}.
\newblock Efficient decision-based black-box adversarial attacks on face recognition.
\newblock In \emph{Proceedings of the IEEE/CVF Conference on Computer Vision and Pattern Recognition}, 7714--7722.

\bibitem[{Dosovitskiy et~al.(2020)Dosovitskiy, Beyer, Kolesnikov, Weissenborn, Zhai, Unterthiner, Dehghani, Minderer, Heigold, Gelly et~al.}]{vit}
Dosovitskiy, A.; Beyer, L.; Kolesnikov, A.; Weissenborn, D.; Zhai, X.; Unterthiner, T.; Dehghani, M.; Minderer, M.; Heigold, G.; Gelly, S.; et~al. 2020.
\newblock An image is worth 16x16 words: Transformers for image recognition at scale.
\newblock \emph{arXiv preprint arXiv:2010.11929}.

\bibitem[{Dziugaite, Ghahramani, and Roy(2016)}]{jpeg}
Dziugaite, G.~K.; Ghahramani, Z.; and Roy, D.~M. 2016.
\newblock A study of the effect of jpg compression on adversarial images.
\newblock \emph{arXiv preprint arXiv:1608.00853}.

\bibitem[{George and Marcel(2019)}]{deeppixbis}
George, A.; and Marcel, S. 2019.
\newblock Deep pixel-wise binary supervision for face presentation attack detection.
\newblock In \emph{2019 International Conference on Biometrics (ICB)}, 1--8. IEEE.

\bibitem[{George et~al.(2019)George, Mostaani, Geissenbuhler, Nikisins, Anjos, and Marcel}]{wmca}
George, A.; Mostaani, Z.; Geissenbuhler, D.; Nikisins, O.; Anjos, A.; and Marcel, S. 2019.
\newblock Biometric face presentation attack detection with multi-channel convolutional neural network.
\newblock \emph{IEEE transactions on information forensics and security}, 15: 42--55.

\bibitem[{Goodfellow, Shlens, and Szegedy(2014)}]{fgsm}
Goodfellow, I.~J.; Shlens, J.; and Szegedy, C. 2014.
\newblock Explaining and harnessing adversarial examples.
\newblock \emph{arXiv preprint arXiv:1412.6572}.

\bibitem[{Gu et~al.(2019)Gu, Yi, Zhu, Yao, and Wang}]{blur}
Gu, S.; Yi, P.; Zhu, T.; Yao, Y.; and Wang, W. 2019.
\newblock Detecting adversarial examples in deep neural networks using normalizing filters.
\newblock \emph{UMBC Student Collection}.

\bibitem[{He et~al.(2022)He, Chen, Xie, Li, Doll{\'a}r, and Girshick}]{mae}
He, K.; Chen, X.; Xie, S.; Li, Y.; Doll{\'a}r, P.; and Girshick, R. 2022.
\newblock Masked autoencoders are scalable vision learners.
\newblock In \emph{Proceedings of the IEEE/CVF conference on computer vision and pattern recognition}, 16000--16009.

\bibitem[{He et~al.(2016)He, Zhang, Ren, and Sun}]{resnet}
He, K.; Zhang, X.; Ren, S.; and Sun, J. 2016.
\newblock Deep residual learning for image recognition.
\newblock In \emph{Proceedings of the IEEE conference on computer vision and pattern recognition}, 770--778.

\bibitem[{Hermans, Beyer, and Leibe(2017)}]{triplet}
Hermans, A.; Beyer, L.; and Leibe, B. 2017.
\newblock In defense of the triplet loss for person re-identification.
\newblock \emph{arXiv preprint arXiv:1703.07737}.

\bibitem[{Heusch et~al.(2020)Heusch, George, Geissb{\"u}hler, Mostaani, and Marcel}]{hqwmca}
Heusch, G.; George, A.; Geissb{\"u}hler, D.; Mostaani, Z.; and Marcel, S. 2020.
\newblock Deep models and shortwave infrared information to detect face presentation attacks.
\newblock \emph{IEEE Transactions on Biometrics, Behavior, and Identity Science}, 2(4): 399--409.

\bibitem[{Ho, Jain, and Abbeel(2020)}]{ddpm}
Ho, J.; Jain, A.; and Abbeel, P. 2020.
\newblock Denoising diffusion probabilistic models.
\newblock \emph{Advances in neural information processing systems}, 33: 6840--6851.

\bibitem[{Huang et~al.(2008)Huang, Mattar, Berg, and Learned-Miller}]{lfw}
Huang, G.~B.; Mattar, M.; Berg, T.; and Learned-Miller, E. 2008.
\newblock Labeled faces in the wild: A database forstudying face recognition in unconstrained environments.
\newblock In \emph{Workshop on faces in'Real-Life'Images: detection, alignment, and recognition}.

\bibitem[{Kim, Kwon, and Ye(2022)}]{diffusionclip}
Kim, G.; Kwon, T.; and Ye, J.~C. 2022.
\newblock Diffusionclip: Text-guided diffusion models for robust image manipulation.
\newblock In \emph{Proceedings of the IEEE/CVF Conference on Computer Vision and Pattern Recognition}, 2426--2435.

\bibitem[{Kingma et~al.(2014)Kingma, Mohamed, Jimenez~Rezende, and Welling}]{vae1}
Kingma, D.~P.; Mohamed, S.; Jimenez~Rezende, D.; and Welling, M. 2014.
\newblock Semi-supervised learning with deep generative models.
\newblock \emph{Advances in neural information processing systems}, 27.

\bibitem[{Komkov and Petiushko(2021)}]{advhat}
Komkov, S.; and Petiushko, A. 2021.
\newblock Advhat: Real-world adversarial attack on arcface face id system.
\newblock In \emph{2020 25th International Conference on Pattern Recognition (ICPR)}, 819--826. IEEE.

\bibitem[{Lin et~al.(2017)Lin, Goyal, Girshick, He, and Doll{\'a}r}]{lin2017focal}
Lin, T.-Y.; Goyal, P.; Girshick, R.; He, K.; and Doll{\'a}r, P. 2017.
\newblock Focal loss for dense object detection.
\newblock In \emph{Proceedings of the IEEE international conference on computer vision}, 2980--2988.

\bibitem[{Liu et~al.(2024)Liu, Xue, Gan, Wan, Liang, Deng, Escalera, and Lei}]{cfplfas}
Liu, A.; Xue, S.; Gan, J.; Wan, J.; Liang, Y.; Deng, J.; Escalera, S.; and Lei, Z. 2024.
\newblock CFPL-FAS: Class Free Prompt Learning for Generalizable Face Anti-spoofing.
\newblock In \emph{Proceedings of the IEEE/CVF Conference on Computer Vision and Pattern Recognition}, 222--232.

\bibitem[{Liu et~al.(2021)Liu, Zhao, Yu, Su, Liu, Kong, Wan, Escalera, Escalante, Lei et~al.}]{hifimask}
Liu, A.; Zhao, C.; Yu, Z.; Su, A.; Liu, X.; Kong, Z.; Wan, J.; Escalera, S.; Escalante, H.~J.; Lei, Z.; et~al. 2021.
\newblock 3d high-fidelity mask face presentation attack detection challenge.
\newblock In \emph{Proceedings of the IEEE/CVF International Conference on Computer Vision}, 814--823.

\bibitem[{Liu, Jourabloo, and Liu(2018)}]{depthnet}
Liu, Y.; Jourabloo, A.; and Liu, X. 2018.
\newblock Learning deep models for face anti-spoofing: Binary or auxiliary supervision.
\newblock In \emph{Proceedings of the IEEE conference on computer vision and pattern recognition}, 389--398.

\bibitem[{Liu et~al.(2019)Liu, Stehouwer, Jourabloo, and Liu}]{liu2019deep}
Liu, Y.; Stehouwer, J.; Jourabloo, A.; and Liu, X. 2019.
\newblock Deep tree learning for zero-shot face anti-spoofing.
\newblock In \emph{Proceedings of the IEEE/CVF Conference on Computer Vision and Pattern Recognition}, 4680--4689.

\bibitem[{Madry et~al.(2017)Madry, Makelov, Schmidt, Tsipras, and Vladu}]{pgd}
Madry, A.; Makelov, A.; Schmidt, L.; Tsipras, D.; and Vladu, A. 2017.
\newblock Towards deep learning models resistant to adversarial attacks.
\newblock \emph{arXiv preprint arXiv:1706.06083}.

\bibitem[{Massoli et~al.(2021)Massoli, Carrara, Amato, and Falchi}]{dfraa}
Massoli, F.~V.; Carrara, F.; Amato, G.; and Falchi, F. 2021.
\newblock Detection of face recognition adversarial attacks.
\newblock \emph{Computer Vision and Image Understanding}, 202: 103103.

\bibitem[{Moitra, Kim, and Panda(2022)}]{est}
Moitra, A.; Kim, Y.; and Panda, P. 2022.
\newblock Adversarial Detection without Model Information.
\newblock \emph{arXiv preprint arXiv:2202.04271}.

\bibitem[{Radford et~al.(2021)Radford, Kim, Hallacy, Ramesh, Goh, Agarwal, Sastry, Askell, Mishkin, Clark et~al.}]{clip}
Radford, A.; Kim, J.~W.; Hallacy, C.; Ramesh, A.; Goh, G.; Agarwal, S.; Sastry, G.; Askell, A.; Mishkin, P.; Clark, J.; et~al. 2021.
\newblock Learning transferable visual models from natural language supervision.
\newblock In \emph{International conference on machine learning}, 8748--8763. PMLR.

\bibitem[{Srivatsan, Naseer, and Nandakumar(2023)}]{srivatsan2023flip}
Srivatsan, K.; Naseer, M.; and Nandakumar, K. 2023.
\newblock FLIP: Cross-domain Face Anti-spoofing with Language Guidance.
\newblock In \emph{Proceedings of the IEEE/CVF International Conference on Computer Vision}, 19685--19696.

\bibitem[{Szegedy et~al.(2017)Szegedy, Ioffe, Vanhoucke, and Alemi}]{inception}
Szegedy, C.; Ioffe, S.; Vanhoucke, V.; and Alemi, A. 2017.
\newblock Inception-v4, inception-resnet and the impact of residual connections on learning.
\newblock In \emph{Proceedings of the AAAI conference on artificial intelligence}, volume~31.

\bibitem[{Tan(2019)}]{efficientnet}
Tan, M. 2019.
\newblock rethinking model scaling for convolutional neural networks. arXiv. 2019 doi: 10.48550.
\newblock \emph{arXiv}.

\bibitem[{Tong et~al.(2021)Tong, Chen, Ni, Cheng, Song, Chen, and Vorobeychik}]{facesec}
Tong, L.; Chen, Z.; Ni, J.; Cheng, W.; Song, D.; Chen, H.; and Vorobeychik, Y. 2021.
\newblock Facesec: A fine-grained robustness evaluation framework for face recognition systems.
\newblock In \emph{Proceedings of the IEEE/CVF Conference on Computer Vision and Pattern Recognition}, 13254--13263.

\bibitem[{Van~der Maaten and Hinton(2008)}]{tsne}
Van~der Maaten, L.; and Hinton, G. 2008.
\newblock Visualizing data using t-SNE.
\newblock \emph{Journal of machine learning research}, 9(11).

\bibitem[{Watson and Al~Moubayed(2021)}]{attack-agnostic}
Watson, M.; and Al~Moubayed, N. 2021.
\newblock Attack-agnostic adversarial detection on medical data using explainable machine learning.
\newblock In \emph{2020 25th International Conference on Pattern Recognition (ICPR)}, 8180--8187. IEEE.

\bibitem[{Wei et~al.(2023)Wei, Mangalam, Huang, Li, Fan, Xu, Wang, Xie, Yuille, and Feichtenhofer}]{diffmae}
Wei, C.; Mangalam, K.; Huang, P.-Y.; Li, Y.; Fan, H.; Xu, H.; Wang, H.; Xie, C.; Yuille, A.; and Feichtenhofer, C. 2023.
\newblock Diffusion Models as Masked Autoencoders.
\newblock \emph{arXiv preprint arXiv:2304.03283}.

\bibitem[{Xie et~al.(2019)Xie, Zhang, Zhou, Bai, Wang, Ren, and Yuille}]{dim}
Xie, C.; Zhang, Z.; Zhou, Y.; Bai, S.; Wang, J.; Ren, Z.; and Yuille, A.~L. 2019.
\newblock Improving transferability of adversarial examples with input diversity.
\newblock In \emph{Proceedings of the IEEE/CVF conference on computer vision and pattern recognition}, 2730--2739.

\bibitem[{Yang et~al.(2023)Yang, Liu, Xu, Wang, Dong, Chen, Su, and Zhu}]{yang2023towards}
Yang, X.; Liu, C.; Xu, L.; Wang, Y.; Dong, Y.; Chen, N.; Su, H.; and Zhu, J. 2023.
\newblock Towards Effective Adversarial Textured 3D Meshes on Physical Face Recognition.
\newblock In \emph{Proceedings of the IEEE/CVF Conference on Computer Vision and Pattern Recognition}, 4119--4128.

\bibitem[{Yu et~al.(2022)Yu, Cai, Li, Yang, Shi, and Kot}]{yu2022bench}
Yu, Z.; Cai, R.; Li, Z.; Yang, W.; Shi, J.; and Kot, A.~C. 2022.
\newblock Benchmarking joint face spoofing and forgery detection with visual and physiological cues.
\newblock \emph{arXiv preprint arXiv:2208.05401}.

\bibitem[{Yu et~al.(2023)Yu, Liu, Zhao, Cheng, Cheng, and Zhao}]{flexibleyu}
Yu, Z.; Liu, A.; Zhao, C.; Cheng, K.~H.; Cheng, X.; and Zhao, G. 2023.
\newblock Flexible-modal face anti-spoofing: A benchmark.
\newblock In \emph{Proceedings of the IEEE/CVF Conference on Computer Vision and Pattern Recognition}, 6345--6350.

\bibitem[{Yu et~al.(2020)Yu, Zhao, Wang, Qin, Su, Li, Zhou, and Zhao}]{CDCN}
Yu, Z.; Zhao, C.; Wang, Z.; Qin, Y.; Su, Z.; Li, X.; Zhou, F.; and Zhao, G. 2020.
\newblock Searching central difference convolutional networks for face anti-spoofing.
\newblock In \emph{Proceedings of the IEEE/CVF Conference on Computer Vision and Pattern Recognition}, 5295--5305.

\bibitem[{Zhang et~al.(2022)Zhang, Liu, Liu, Ramachandra, and Busch}]{frt}
Zhang, W.; Liu, H.; Liu, F.; Ramachandra, R.; and Busch, C. 2022.
\newblock Effective Presentation Attack Detection Driven by Face Related Task.
\newblock In \emph{European Conference on Computer Vision}, 408--423. Springer.

\end{thebibliography}

\end{document}